%% file: main_arxiv.tex
\documentclass[10pt,twocolumn,letterpaper]{article}

\usepackage{iccv}
\usepackage{times}
\usepackage{epsfig}
\usepackage{graphicx}
\usepackage{amsmath}
\usepackage{amssymb}
\usepackage{multirow}
\usepackage{xcolor,colortbl}
\usepackage{algorithm}
\usepackage[noend]{algpseudocode}
\usepackage{tabularx}
\usepackage{booktabs}
\usepackage{enumitem}
\usepackage{subfig}
\usepackage{xcolor}
\usepackage{xspace}
\makeatletter
\@namedef{ver@everyshi.sty}{}
\makeatother
\usepackage{tikz}

\graphicspath{{./figures/}}

\def\BA2{\textit{BA}$^2$}

\def\ker{\boldsymbol{K}}

\DeclareMathOperator*{\argmin}{\arg\min}

\newcommand{\copyrighttext}{%
  \footnotesize%
  \copyright\xspace2019 IEEE. Personal use of this material is permitted. Permission from IEEE must be obtained for all other uses, in any current or future media, including reprinting/republishing this material for advertising or promotional purposes, creating new collective works, for resale or redistribution to servers or lists, or reuse of any copyrighted component of this work in other works.}
\newcommand{\copyrightnotice}{%
\begin{tikzpicture}[remember picture,overlay]
{\setlength{\fboxsep}{0pt}%
 \setlength{\fboxrule}{0pt}%
 \node[anchor=south,yshift=10pt,xshift=-7pt] at (current page.south) {\fbox{\parbox{\dimexpr\textwidth-\fboxsep-\fboxrule\relax}{\copyrighttext}}};
}
\end{tikzpicture}%
}

\usepackage[breaklinks=true,bookmarks=false]{hyperref}
\hypersetup{
    colorlinks,
    linkcolor={red},
    citecolor={blue},
    urlcolor={pink}
}

\iccvfinalcopy % *** Uncomment this line for the final submission

\def\iccvPaperID{4541} % *** Enter the ICCV Paper ID here

\ificcvfinal\fi

\begin{document}

%%%%%%%%% TITLE
\input{sections/0_title}

\maketitle
\ificcvfinal\thispagestyle{empty}\fi
\copyrightnotice
\vspace{-10pt}

%%%%%%%%% ABSTRACT
\input{sections/1_abstract}

%%%%%%%%% BODY TEXT
\input{sections/2_introduction}

\input{sections/3_related}

\input{sections/4_method}

\input{sections/5_experiments}
\input{sections/6_conclusions}

{\small
\bibliographystyle{ieee_fullname}
\bibliography{budget_conv}
}

\clearpage
\normalsize

\appendix
\renewcommand\thetable{\Alph{section}.\arabic{table}}
\setcounter{table}{0} 

\section*{Appendix}
Here, we first provide more details about the training and evaluation protocols used in our experiments (Appendix~\ref{sec:details}). Then, in Appendix~\ref{sec:denseNet}, we report additional experiments on the ImageNet-to-Sketch benchmark.

\input{sections/7_supplementary.tex}

\end{document}

%% file: sections/0_title.tex
\title{Budget-Aware Adapters for Multi-Domain Learning}

\author{
    Rodrigo Berriel$^1$\thanks{Work done when with MHUG (DISI, University of Trento).} \quad
    St\'ephane Lathuili\`ere$^2$ \quad 
    Moin Nabi$^3$ \quad 
    Tassilo Klein$^3$ \and 
    Thiago Oliveira-Santos$^1$ \quad 
    Nicu Sebe$^2$ \quad
    Elisa Ricci$^{2,4}$
    \\
    $^1 $LCAD, UFES \quad 
    $^2 $DISI, University of Trento \quad
    $^3 $SAP ML Research \quad 
    $^4 $Fondazione Bruno Kessler
    \\
    {\tt\small berriel@lcad.inf.ufes.br}
}

%% file: sections/1_abstract.tex
\begin{abstract}

Multi-Domain Learning (MDL) refers to the problem of learning a set of models derived from a common deep architecture, each one specialized to perform a task in a certain domain (e.g., photos, sketches, paintings). This paper tackles MDL with a particular interest in obtaining domain-specific models with an adjustable budget in terms of the number of network parameters and computational complexity. Our intuition is that, as in real applications the number of domains and tasks can be very large, an effective MDL approach should not only focus on accuracy but also on having as few parameters as possible. To implement this idea we derive specialized deep models for each domain by adapting a pre-trained architecture but, differently from other methods, we propose a novel strategy to automatically adjust the computational complexity of the network. To this aim, we introduce \emph{Budget-Aware Adapters} that select the most relevant feature channels to better handle data from a novel domain. Some constraints on the number of active switches are imposed in order to obtain a network respecting the desired complexity budget. Experimentally, we show that our approach leads to recognition accuracy competitive with state-of-the-art approaches but with much lighter networks both in terms of storage and computation.

\end{abstract}
\vspace{-0.25cm}

%% file: sections/2_introduction.tex
\section{Introduction}

\begin{figure}[t]\centering
\includegraphics[width=0.92\linewidth]{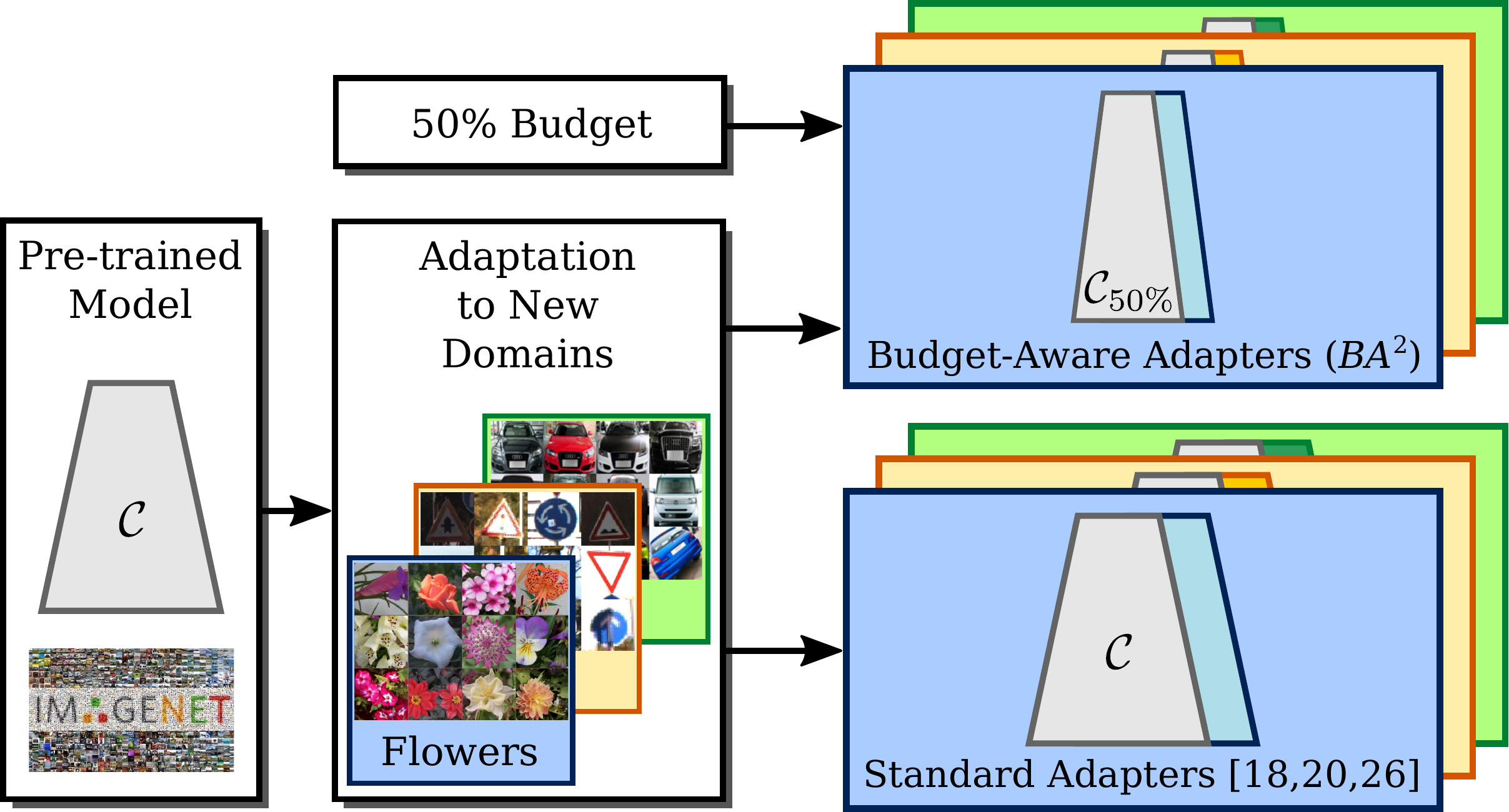}
\caption{In Multi-Domain Learning, a pre-trained model is usually adapted to solve new tasks in new domains. When using standard approaches, the complexity $\mathcal{C}$ of the domain-specific models is dependent on the pre-trained model complexity. In this work we propose a novel approach to learn specialized models while imposing budget constraints in terms of the number of parameters for each new domain. %  the model complexity can be adjusted to obtain the best trade-off between accuracy and model complexity. 
%\steph{change references in teaser and fig 4}
}
\vspace{-0.3cm}
\label{fig:pipeline}
\end{figure}

Deep learning methods have brought revolutionary advances in computer vision, setting the state of the art in many tasks such as object recognition \cite{he2016deep,krizhevsky2012imagenet}, detection \cite{girshick2014rich}, semantic segmentation \cite{chen2018deeplab}, depth estimation \cite{xu2017multi}, and many more. Despite these progresses, a major drawback with deep architectures is that when a novel task is addressed typically a new model is required. However, in many situations it may be reasonable to learn models which perform well on data from different domains. This problem, referred as Multi-Domain Learning (MDL) and originally proposed in \cite{rebuffi2017learning}, has received considerable attention lately \cite{mallya2018piggyback,mancini2018eccvw,rebuffi2018efficient}. An example of MDL is the problem of image classification when the data belong to several domains (e.g., natural images, paintings, sketches, etc.) and the categories in the different domains do not overlap.% \berriel{This is a valid example indeed, but they do overlap on the Visual Decathlon (e.g., ImageNet and CIFAR-100).}

Previous MDL approaches \cite{mallya2018piggyback,mancini2018eccvw,rebuffi2017learning,rebuffi2018efficient} utilize a common backbone architecture (i.e., a pre-trained model) and learn a {limited set} of {domain-specific parameters}. This strategy is advantageous with respect to building several independent classifiers, as it guarantees a significant saving in terms of memory. Furthermore, it naturally deals with the \textit{catastrophic forgetting} issue, as when a new domain is considered the knowledge on the previously learned ones is retained. Existing approaches mostly differ from the way domain-specific parameters are designed and integrated within the backbone architecture. For instance, binary masks are employed in \cite{mallya2018piggyback,mancini2018eccvw} in order to select the parameters of the main network that are useful for a given task. Differently, in \cite{rebuffi2017learning,rebuffi2018efficient} domain-specific residual blocks are embedded in the original deep architecture. While the different approaches are typically compared in terms of classification accuracy, their computational and memory requirements are not taken into account.

In this paper we argue that an optimal MDL algorithm should not only achieve high recognition accuracy on all the different domains but should also permit to keep the number of parameters as low as possible. In fact, %ideally MDL methods should enable the learning of classification models for an arbitrarily large number of tasks. Therefore, 
as in real world applications the number of domains and tasks can be very large, it is desirable to limit the models' complexity (both in terms of memory and computation). Furthermore, it is very reasonable to assume that different domains and tasks may correspond to a different degree of difficulty (e.g., recognizing digits is usually easier than classifying flowers) and may require different models: small networks should be used for easy tasks, while models with a large number of parameters should be employed for difficult ones. %\berriel{I wonder if we should reverse the order: easy tasks ``require'' small networks, while difficult ones may require more parameters.}

Following these ideas, we propose the first MDL approach which derives a set of domain-specific classifiers from a common backbone deep architecture under a budget constraint, where the budget is specified by a user and is expressed as the number of network parameters (see Figure \ref{fig:pipeline}).
This idea is realized by designing a new network module, called \emph{Budget-Aware Adapters} (\BA2), which embeds switch variables that can select the feature channels relevant to a domain. By dropping feature channels in each convolutional layer, \BA2 both adapt the image representation of the network and reduce the computational complexity. Furthermore, we propose a constrained optimization problem formulation in order to train domain-specific classifiers that respect budget constraints provided by the user. The proposed approach has been evaluated on two publicly available benchmarks, the ten datasets of the Visual Decathlon Challenge~\cite{rebuffi2017learning} and the six-dataset benchmark proposed in~\cite{mallya2018piggyback}. Our results show that the proposed method is competitive with state-of-the-art baselines and requires much less storage and computational resources.

%\eli{contributions???- we may want to skip}

%\eli{to set it aside from multi-task learning - where different tasks are to be performed on the same domain. }

%% file: sections/3_related.tex
\section{Related Work}

\textbf{Multi-domain Learning.} The problem of adapting deep architectures to novel tasks and domains has been extensively studied in the past. Earlier works considered simple strategies, such as fine-tuning existing pre-trained models, with the drawback of incurring to \textit{catastrophic forgetting} and of requiring the storage of multiple specialized models. More recent studies address the problem proposing methods for extending the capabilities of existing deep architectures by adding few task-specific parameters.
In this way, as the parameters of the original network are left untouched, the catastrophic forgetting issue is naturally circumvented. For instance, Rebuffi \etal~\cite{rebuffi2017learning} introduced residual adapters, i.e., a novel design for residual blocks that embed task-specific components. In a subsequent work \cite{rebuffi2018efficient}, they proposed an improved architecture where the topology of the adapters is parallel rather than series. Rosenfeld \etal~\cite{rosenfeld2017incremental} %presented Deep Adaptation Modules that 
employed controller modules to constrain newly learned parameters to be linear combinations of existing ones. Weight-based pruning has been considered in \cite{mallya2017packnet} to adapt a single neural network to multiple tasks. Aiming at decreasing the overhead in terms of storage, more recent works proposed to adopt binary masks \cite{mallya2018piggyback,mancini2018eccvw} as task-specific parameters. In particular, while in \cite{mallya2018piggyback} simple multiplicative binary masks are used to indicate which parameters are and which are not useful for a new task, \cite{mancini2018eccvw} proposes a more general formulation considering affine transformations. Guo \etal~\cite{guo2019spottune} proposed an adaptive fine-tuning method and derive specialized classifiers by fine-tuning certain layers according to a given target image.

While these works considered a supervised learning setting, the idea of learning task-specific parameters has also been considered in reinforcement learning. For instance Rusu \textit{et al.} \cite{rusu2016progressive} proposed an approach where each novel task is addressed by adding a side branch to the main network. While our approach also aims at developing architectures which adapts a pretrained model to novel tasks, we target for the first time the problem of automatically adjusting the complexity of the task-specific models.

\textbf{Incremental and Life-long learning.}
%The keen interest on incremental and life-long learning methods dates back to the pre-convnet era, with shallow learning approaches ranging from large margin classifiers \cite{KuzborskijOC13,KuzborskijOC17} to non-parametric methods \cite{MensinkVPC13,RistinGGG16}. 
%WORKS ON MTL/IL:
In the last few years several works have addressed the problem of incremental \cite{BendaleB16,RebuffiKSL17} and life-long learning  \cite{aljundi2018memory,kirkpatrick2017overcoming,li2017learning}, considering different strategies to avoid catastrophic forgetting. %A major risk when training a neural network on a novel task is to deteriorate the performances of the network on old tasks, discarding previous knowledge. This phenomenon is called \textit{catastrophic forgetting} \cite{mccloskey1989catastrophic,french1999catastrophic,goodfellow2013empirical}.
For instance, Li and Hoeim~\cite{li2017learning} proposed to adopt knowledge distillation to ensure that the model adapted to the new tasks is also effective for the old ones. Kirkpatrick \etal~\cite{kirkpatrick2017overcoming} demonstrated that a good strategy to avoid forgetting on the old tasks is to selectively slow down learning on the weights important for those tasks. In \cite{aljundi2018memory} Aljundi \etal presented Memory Aware Synapses, where the idea is to estimate the importance weights for the network parameters in an unsupervised manner in order to allow adaptation to unlabeled data stream. However, while these works are interested in learning over multiple tasks in sequence, in this paper we focus on a different problem, i.e., re-configuring an existing architecture under some resource constraints.

%While these approaches are optimal in terms of the required parameters, i.e. they maintain the same number of parameters of the original network, they limit the catastrophic forgetting problem to the expenses of a lower performance on both old and new tasks. 

\textbf{Adaptive and Resource-aware Networks.} The problem of designing deep architectures which allow an adaptive accuracy-efficiency trade-off directly at runtime has been recently addressed in the research community. For instance, Wu \etal~\cite{wu2018blockdrop} proposed BlockDrop, an approach that learns to dynamically choose which layers of a Residual Network to drop at test time to reduce the computational cost while retaining the prediction accuracy. Wang \etal~\cite{wang2018skipnet} introduced novel gating functions to automatically define at test time the computational graph based on the current network input. Slimmable Networks have been introduced in \cite{yu2018slimmable} with the purpose of adjusting the network width according to resource constraints. While our approach is inspired by these methods, in this paper we show that the idea of dynamically adjusting the network according to resource constraints is especially beneficial in the multi-domain setting.

%% file: sections/4_method.tex
\section{Budget-Aware Adapters for MDL }
\label{sec:method}

In Multi-Domain Learning (MDL), the goal is to learn a single model that can work for diverse visual domains, such as pictures from the web, medical images, paintings, etc. Importantly, when the visual domains are very different, the model has to adapt its image representation. To address MDL, we follow the common approach~\cite{mallya2018piggyback,rebuffi2017learning} that consists in learning Convolutional Neural Networks (ConvNets) that share the vast majority of their parameters but employ a very limited number of additional parameters specifically trained for each domain. %This strategy has been considered in~\cite{mallya2018piggyback,rebuffi2017learning} where a pre-trained network is employed and adapted to multiple domain.

Formally, we consider an arbitrary pre-trained ConvNet $\Psi_0(\cdot; \theta_0):\mathcal{X} \to\mathcal{Y}_0$ with parameters $\theta_0$ that assigns class labels in ${\cal Y}_0$ to elements of an input space $\cal X$ (e.g., images). Our goal is to learn for each domain $d\in\{1,\ldots, D\}$, 
a classifier $\Psi_d(\cdot;\theta_0,\theta_a^d):\mathcal{X}\to \mathcal{Y}_d$ with a possibly different output space ${\cal Y}_d$ that shares the vast majority of its parameters $\theta_0$
but exploits additional domain-specific parameters $\theta_a^d$ to adapt $\Psi_d$ to the domain $d$.

In this paper, we claim that an effective approach for MDL should require a low number of domain-specific parameters. In other words, the cardinality of each $\theta_a^d$ parameter set should be negligible with respect to the cardinality of $\theta_0$. 
In addition, we argue that one major drawback of previous MDL methods is that the network computational complexity directly ensues from the initial pre-trained network $\Psi_0$. 
More precisely, the networks $\Psi_d$ for the new domains usually have computational complexities at best equal to the one of the initial pre-trained network. Moreover, such models lack flexibility for deployment since the user cannot adjust the computational complexity of $\Psi_d$ depending on its needs or on hardware constraints.

To address this issue, we introduce novel modules, the \emph{Budget-Aware Adapters} (\BA2) that are designed both for enabling a pre-trained model to handle a new domain and for controlling the network complexity. The key idea behind \BA2 is that the parameters $\theta_a^d$ control the use of the convolution operations parametrized by $\theta_0$. Therefore, \BA2 can learn to drop parts of the computational graph of $\Psi_0$ and parts of the parameters $\theta_0$ resulting in a model $\Psi_d$ with a lower computational complexity and fewer parameters to load at inference time.
In the following, we first describe the proposed \emph{Budget-Aware Adapters} (Subsection~\ref{sec:budget-conv}) and then present the training procedure we introduced to learn domain-specific models with budget constraints % for Multi-Domain Learning is detailed in 
(Subsection~\ref{sec:budget-conv-training}).

\subsection{Adapting Convolutions with \BA2}
\label{sec:budget-conv}

\begin{figure}[t]
	\begin{center}
		%\fbox{\rule{0pt}{1in} \rule{0.9\linewidth}{0pt}}
		\includegraphics[width=0.99\linewidth]{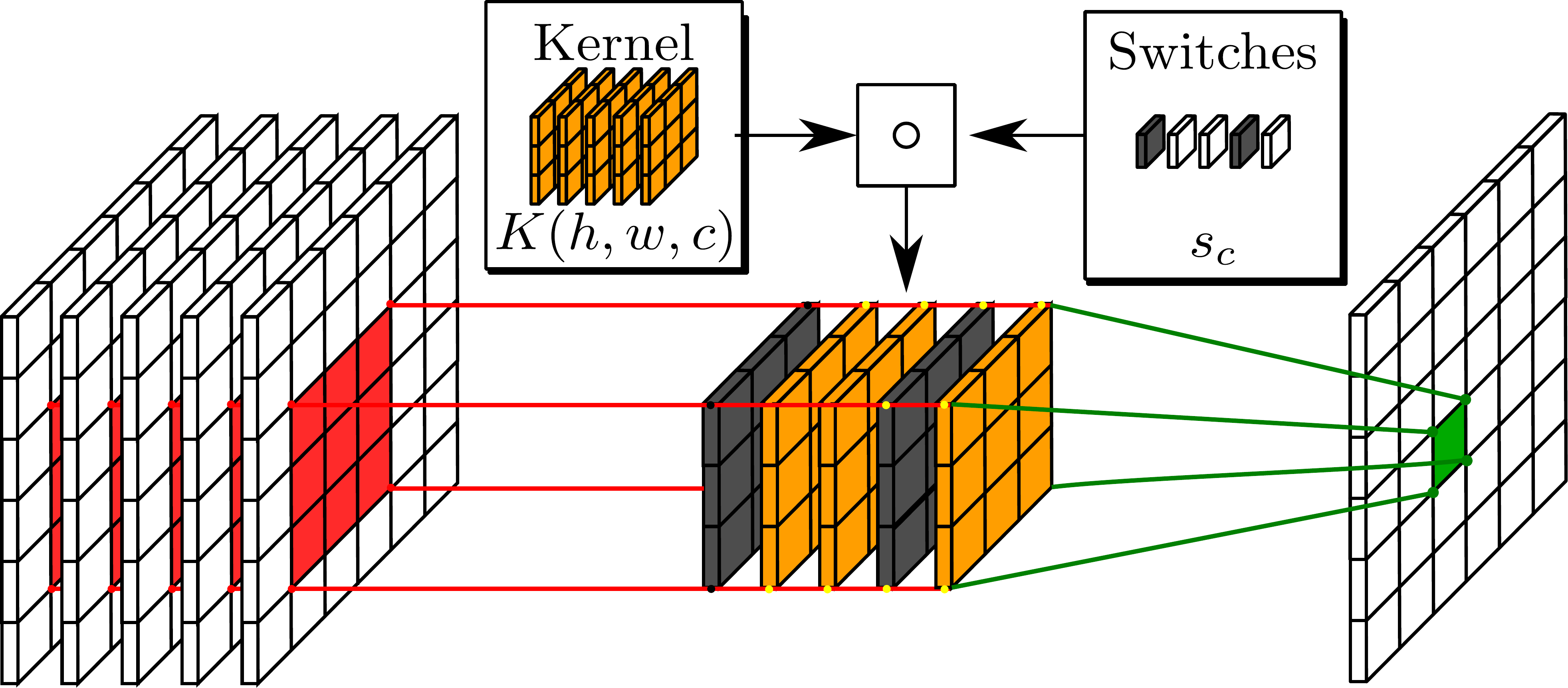}
	\end{center}
	\caption{Budget-Aware Adapters (\BA2): a switch vector controls  the activation of convolution channels in order to both adapt the network to a new domain and adjust its computational complexity. Dark grey arrays represents channels that are ``turned off'' by the switches.}
	\label{fig:mask}
\end{figure}

We now describe our \emph{Budget-Aware Adapters} illustrated in Figure~\ref{fig:mask}. Since, \BA2 acts on the elementary convolution operation, it can be employed in any ConvNet but, for the sake of notation simplicity, we consider the case of 2D convolutions. Let $\ker \in \mathbb{R}^{2K_H+1 \times 2K_W+1 \times C}$ be a kernel of a standard convolutional layer of $\Psi_0$. Here $2K_H+1$ and $2K_W+1$ denote the kernel size and $C$ the number of input channels. Note that $\ker$ is a subset of the parameters $\theta_0$ introduced in Section~\ref{sec:method}.
Considering a standard 2D convolution, an input feature map $\boldsymbol{I} \in \mathbb{R}^{H \times W \times C}$ and an activation function $g$, the output value at the location $(i,j)\in [1..H] \times [1..W]$ is given by:
\begin{align}
  \boldsymbol{x}(i,j) = g\Big(\sum_{c=1}^{C}\phi_c(i,j)\Big),
\end{align}
%% \vspace{-0.5cm}
  where $\phi_c$ is given by:
\begin{align}
  \phi_c=\sum_{h=-K_h}^{K_h}\sum_{w=-K_w}^{K_w}\ker
  (h,w,c)\boldsymbol{I}(i-h,j-w,c).\label{eq:phi}
\end{align}
For the sake of simplicity, the kernel parameter tensor $\ker$ is indexed from $-K_h$ to $K_h$ and from $-K_w$ to $K_w$. 
When learning a new domain $d$, we propose to adapt the convolution by controlling the use of each channel of the convolution layer. To this aim, we introduce an additional binary switch vector $s\in\{0,1\}^C$. This vector $s$ is a subset of the $\theta_a^d$ introduced in Section~\ref{sec:method}. As shown in Figure~\ref{sec:budget-conv}, each switch value is used for an entire channel. As a consequence, \BA2 results in a limited number of additional parameters. Formally, the output of the adapted convolution at location $(i,j)$ is given by: 
\begin{align}
	\boldsymbol{x}(i,j) = g\Big(\sum_{c=1}^{C}s_c\phi_c(i,j)\Big)\label{eq:ba2}
\end{align}
Note that, when $s_c=0$, the tensor $\phi_c$ in Equation~\eqref{eq:ba2} does not need to be computed. In this context, by adjusting the proportion of zeros in  $s_c$, we can control the computational complexity of the convolution layer. Furthermore, in Equation~\eqref{eq:phi}, when $\phi_c$ is not computed, the kernel weights values $\ker(h,w,c)$ can be removed from the computational graph and do not need to be stored. Therefore, in scenarios where the parameters $\ker$ of the initial networks are not further used, these $\ker(h,w,c)$ weights can be dropped resulting in a lower number of parameters to store. Thus, $s_c$ can also control the number of parameters of the new domain networks.
%% The $s_c$ is a binary value that acts as a switch that can mask out an input channel. Interestingly, the dimension of $s_c$ is much smaller than the dimension of $\ker$. Therefore, Budget adapter requires a very limited number of additional parameters for learning a new domain.    

In order to obtain a model that can be trained via Stochastic Gradient Descent, we follow \cite{kaiser2018fast,mallya2018piggyback} and obtain binary values using a threshold function $\tau$:
\begin{equation}
    s_c=\tau(\tilde{s}_c)=
    \left\{ \begin{array}{ll}
        0 & \tilde{s}_c\leq 0.0 \\
        1 &\text{otherwise}
    \end{array} \right.
\end{equation}
where $\tilde{s}_c\in \mathbb{R}$ are continuous scalar parameters.
Similarly to \cite{kaiser2018fast,mallya2018piggyback}, during backward propagation, the $\tau$ function is replaced by the identity function to be able to back-propagate the error and update $\tilde{s}_c$. Even though we learn  $\tilde{s}_c$ at training time, we only need to store the binary $s_c$ values to use at testing time, leading to a small storage requirement (1-bit per $s_c$). Compared to other multi-domain methods that generally use additional 32-bit floating-point numbers \cite{rebuffi2017learning,rebuffi2018efficient,rosenfeld2017incremental}, \BA2 results in a much lighter storage. 

The proposed \BA2 have four main features:

%% Universal parametric network family
\noindent \textbf{Adapting image representation:} 
In \BA2, the $\phi_c$ features can be interpreted as a filter bank and the switch vectors can be understood as a filter selector. Depending on the domain, different switch values can be employed to select features relevant for the considered domain. %In other words, our adapters can be employed to define a parametric network family where the kernel parameters $\ker$ are shared but switches are used to re-parameterize the network specifically to each domain.

\noindent \textbf{Low computational complexity:}
After training, all the tensors $\{\phi_c \mid s_c=0\}$ can be removed of the computational graph, resulting in a lower computational complexity. More precisely, the computational complexity is proportional to:
\begin{align}
    \mathcal{C}=\frac{1}{C}\sum_{c=1}^{C}s_c.
    \label{comp}
\end{align}
Note that, the uncomputed operations are grouped in channels allowing fast GPU implementation. %\berriel{Should I try to find a reference to back this up? Fast Sparse Conv GPU implementation.}

\noindent \textbf{Lower storage:}
First, the number of additional parameters is rather small compared to the number of kernel parameters of the base network.
Second, at testing time, the additional switch parameters can be stored with a binary representation to obtain a lightweight storage (1-bit per kernel channel).
Finally, the weight values $\{\ker(h,w,c) \mid s_c=0\}$ can be dropped, obtaining models with fewer parameters for the new domains. Again, the number of parameters is proportional to $\mathcal{C}$ in Equation~\eqref{comp}.

\noindent \textbf{Low Memory footprint:}
Reducing the computational complexity, does not necessarily reduces the memory footprint at testing time. In order to properly reduce the memory footprint, one needs to reduce the memory requirements of all operations across the computational graph, as stated in \cite{mobilenet2018cvpr}. Given that \BA2 works on the level of the convolution operation, it can also control the memory footprint.

\input{tables/decathlon.tex}

\subsection{Training Budget-aware adapters for MDL}
\label{sec:budget-conv-training}
%% (both computational and memory), i.e., to use the least amount of parameters required to achieve maximum performance

We now detail how \BA2 is used for MDL.
As explained in Subsection~\ref{sec:budget-conv}, we follow a strategy of adapting a pre-trained model to novel domains. Therefore, when learning a new domain, we consider that $\theta_0$ is provided and we keep it fixed for the whole training procedure.% \berriel{Should we say that as most related works, we also assume that $\theta_0$ was trained on a very generic domain, such as ImageNet?}
As a consequence the learning procedure can be subdivided in independent training for each domain resulting in simpler training procedure.
Note that, similarly to \cite{mallya2018piggyback,mancini2018eccvw,rebuffi2017learning}, we use batch-normalization parameters specific for each domain. Furthermore, as shown in \cite{yu2018slimmable}, using different number of channels leads to different feature mean and variance and, as a consequence, sharing Batch Normalization layers performs poorly. Therefore, we use different batch-normalization layers for each budget. Note that, since the number of parameters in a batch-normalization layer is much lower than in convolution layer, this solution does not increase significantly the number of additional parameters with respect to the size of $\theta_0$.
%\berriel{I think this (BN-related stuff) should be in the end. But the next sentence depends on it :(}

Following the notations introduced in Section~\ref{sec:method}, $\theta_a^d$ now denotes the set of all the switch values $s_c$ and the additional batch-normalization parameters.
Considering a new domain $d$, $\Psi_d$ is trained using a loss $\mathcal{L}$. In the case of classification, we employ the cross-entropy loss for all the domains. %\berriel{We only perform classification now. Should we remove the ``In the case of...''?}
In the context of \BA2, we aim at training $\Psi_d$  with budget constraints. Formally, we formulate the optimization problem as follows: we minimize $\mathcal{L}$ with respect to the \BA2 parameters $\theta_a^d$ such that the network complexity satisfies a target budget $\beta\in[0,1]$. For each new domain, we obtain the following constrained optimization problem:
\begin{align}
  \theta_a^{d*} =& \argmin_{\theta_a^d} \mathcal{L}(\theta_0,\theta_a^d) \label{opt} \\
   &s.t.~\bar{\theta_a^d} \leq \beta \label{eq:opc}
\end{align}
where $\bar{\theta_a^d}$ denotes the mean value of the switches in $\theta_a^d$.
From Equation~\eqref{opt}, we construct the generalized Lagrange function and the associated optimization problem:
\begin{align}
  \theta_a^{d*} =& \argmin_{\theta_a^d}
 \big[\mathcal{L}(\theta_0,\theta_a^d)+ \max_{\lambda \geq 0}(\lambda(\bar{\theta_a^d} - \beta))\big]. \label{KKT}
\end{align}
The $\lambda$ is known as the Karush-Kuhn-Tucker (KKT) multiplier. Equation~\eqref{KKT} is optimized via stochastic gradient descent (SGD). When the budget constrained is respected, $\lambda=0$ and Equation~\eqref{KKT} corresponds to $\mathcal{L}$ minimization. When the constraint is not satisfied, in addition to $\mathcal{L}$ minimization, the SGD steps also lead to an increase of $\lambda$ which in turn increases the impact of the budget constraint on $\mathcal{L}$.

In order to obtained networks with different budgets, training is performed independently for each $\beta$ value.
When $\beta$ is set to 1, the constraint in Equation~\eqref{opt} is satisfied for any $\theta_a^d$. Therefore, the problem consists in a loss minimization problem over the parametric network family defined by $\theta_0$. This scenario corresponds to a standard multi-domain scenario without considering budget as in \cite{mallya2018piggyback,mancini2018eccvw,rebuffi2017learning,rosenfeld2017incremental}.
When $\beta<1$, we combine both re-parametrization and budget-adjustable abilities of \emph{Budget-Aware Adapters}. In this case, the goal is to obtain the best performing model that respects the budget constrain. It is important to note that the actual complexity of the network, after training, can be lower than the one defined by the user, including the $\beta=1$ case.

Note that in Equation~\eqref{opt}, the budget constraint is formulated as a constraint on the total network complexity. In practice, it can be preferable to constrain each \BA2 to satisfy independent budget constraints in order to both spread computation over the layers and obtain a lower memory footprint. In this case, KKT multipliers are added in Equation \eqref{KKT} for each convolution layer.

%% file: tables/decathlon.tex
\begin{table*}[ht]
       \centering
    \resizebox{0.97\textwidth}{!}{%
\begin{tabular}{ l | c || c  c  c  c  c  c  c  c  c  c | c || c} 
Method&Params&ImNet&Airc. &C100&DPed&DTD&GTSR&Flwr.&Oglt.&SVHN&UCF&Mean& $S$-Score\\\toprule
Feature \cite{rebuffi2017learning}&\bf 1&59.7	&23.3	&63.1	&80.3	&45.4	&68.2	&73.7	&58.8	&43.5	&26.8	&54.3&544\\
Finetune \cite{rebuffi2017learning}&10&59.9	&60.3	&82.1	&92.8	&55.5	&97.5	&81.4	&87.7	&96.6	&51.2	&76.5&2500\\
SpotTune \cite{guo2019spottune} & 11 &60.3 & 63.9 & 80.5 & 96.5  & 57.13  & \textbf{99.5} &  85.22 & 88.8 & 96.7 & \textbf{ 52.3 }& \textbf{ 78.1} & \textbf{3612}\\\midrule
RA\cite{rebuffi2017learning}&2&59.7	&56.7	&\underline{81.2}	&93.9	&50.9	&97.1	&66.2	&\underline{89.6}	&96.1	&47.5	&73.9&2118\\
DAM \cite{rosenfeld2017incremental}&2.17&57.7	&64.1	&80.1	&91.3	&56.5	&98.5	&\underline{86.1}	&\textbf{89.7}	&96.8	&49.4	&\underline{77.0}&2851\\
PA \cite{rebuffi2018efficient}&2&\underline{60.3}&\underline{64.2}&\textbf{81.9}&94.7&58.8&99.4&84.7&89.2&96.5&\underline{50.9}&\textbf{78.1}&3412\\
PB \cite{mallya2018piggyback}&1.28&57.7	&\textbf{65.3}	&79.9	&\textbf{97.0}	&57.5	&97.3	&79.1	&87.6	&\textbf{97.2}	&47.5	&76.6&2838\\
WTPB \cite{mancini2018eccvw} &1.29&\textbf{60.8}      &52.8 &\textbf{82.0}    &\underline{96.2} &58.7 &99.2 &\textbf{88.2}    &89.2 &96.8 &48.6 &77.2&\underline{3497}\\\midrule
\BA2 (Ours) ($\beta=1.00$) & \underline{1.03} & 56.9 & 49.9 & 78.1 & 95.5 & 55.1 & \underline{99.4} & \underline{86.1} & 88.7 & \underline{96.9} & 50.2 & 75.7  & 3199\\
%\BA2 (Ours) ($\beta=0.75$) & 1.03 & 56.9 & 47.5 & 77.9 & 94.5 & 54.9 & 99.1 & 83.7 & 88.8 & 96.7 & 47.6 & 74.8 & 2876\\
%\BA2 (Ours) ($\beta=0.50$) & 1.03 & 56.9 & 45.5 & 77.3 & 94.5 & 51.6 & 99.2 & 83.3 & 88.7 & 96.9 & 47.5 & 74.1 & 2856\\
%\BA2 (Ours) ($\beta=0.25$) & 1.03 & 56.9 & 43.2 & 71.6 & 95.1 & 50.4 & 99.2 & 79.5 & 88.8 & 96.6 & 43.5 & %72.5 & 2589\\
%\hline \BA2 (Ours) KKT ($\beta=1.00$) & 1.03 & 56.9 & 49.8 & 78.5 & 94.5 & 57.1 & 99.4 & 84.8 & 89.0 & 97.1 & 48.7 & 75.6  & 3148\\
\BA2 (Ours) ($\beta=0.75$) & \underline{1.03} & 56.9 & 47.0 & 78.4 & 95.3 & 55.0 & 99.2 & 85.6 & 88.8 & 96.8 & 48.7 & 75.2  & 3063\\
\BA2 (Ours) ($\beta=0.50$) & \underline{1.03} & 56.9 & 45.7 & 76.6 & 95.0 & 55.2 & 99.4 & 83.3 & 88.9 & 96.9 & 46.8 & 74.5  & 2999\\
\BA2 (Ours) ($\beta=0.25$) & \underline{1.03} & 56.9 & 42.2 & 71.0 & 93.4 & 52.4 & 99.1 & 82.0 & 88.5 & 96.9 & 43.9 & 72.6  & 2538\\
\end{tabular}%
	} 
	\caption{Results in terms of accuracy and $S$-Score, for the Visual Decathlon Challenge. Best model in bold, second best underlined. %\berriel{DAN changed to DAM (Network to Module). Looks like most methods only add parameters at training time, such that they can be "fused" at test time.}
	}
    \label{tab:vdc-accuracy}
\end{table*}

%% file: sections/5_experiments.tex
\section{Experimental Results}
In this section we present the experimental methodology and metrics used to evaluate our approach. Moreover, we report the results and comparisons with state of the art MDL approaches (Subsection \ref{exp:MDL}). In addition, we also conduct further experiments on the usual single-domain setting and demonstrate the effectiveness of \BA2 (Subsection \ref{exp:baa}) in reducing complexity while learning accurate recognition models.

\subsection{Multi-Domain Learning}
\label{exp:MDL}

\paragraph{Datasets.}
In order to evaluate our MDL approach, we adopt two different benchmarks. 
We first consider the Visual Decathlon Challenge~\cite{rebuffi2017learning}. The purpose of this challenge is to compare methods for MDL over 10 different classification tasks: ImageNet~\cite{russakovsky2015imagenet}, CIFAR-100~\cite{krizhevsky2009learning}, Aircraft~\cite{maji2013fine}, Daimler pedestrian (DPed)~\cite{munder2006experimental}, Describable Textures (DTD)~\cite{cimpoi2014describing}, German Traffic Signs (GTSR)~\cite{stallkamp2012man}, Omniglot \cite{lake2015human}, SVHN~\cite{netzer2011reading}, UCF101 Dynamic Images~\cite{bilen2016dynamic,soomro2012ucf101} and VGG-Flowers~\cite{nilsback2008automated}. For more details about the challenge, please refer to~\cite{rebuffi2017learning}. 

As for the second benchmark, we follow previous works~\cite{mallya2018piggyback,mancini2018eccvw} and consider the union of six different datasets: ImageNet~\cite{russakovsky2015imagenet}, VGG-Flowers~\cite{nilsback2008automated}, Stanford Cars~\cite{krause20133d}, Caltech-UCSD Birds (CUBS)~\cite{wah2011caltech}, Sketches~\cite{eitz2012humans}, and WikiArt~\cite{saleh2015large}. % MASSI BULLSHIT THEN VGG-Flowers \cite{nilsback2008automated} is a dataset of fine-grained recognition containing images of 102  categories, corresponding to different kind of flowers. There are 2'040 images for training and 6'149 for testing. Stanford Cars \cite{krause20133d} contains images of 196 different types of cars with approximately 8 thousand images for training and 8 thousands for testing. Caltech-UCSD Birds \cite{wah2011caltech} is another dataset of fine-grained recognition containing images of 200 different species of birds, with approximately 6 thousands images for training and 6 thousands for testing. Sketches \cite{eitz2012humans} is a dataset composed of 20 thousands sketch drawings, 16 thousands for training and 4 thousands for testing. It contains images of 250 different objects in their sketched representations. WikiArt \cite{saleh2015large} contains painting from 195 different artists. The dataset has 42129 images for training and 10628 images for testing.
These datasets are very heterogeneous, comprising a wide range of the categories (e.g., cars \cite{krause20133d} vs birds \cite{wah2011caltech}) and a large variety of image appearance (i.e., natural images~\cite{russakovsky2015imagenet}, art paintings~\cite{saleh2015large}, sketches~\cite{eitz2012humans}).

%, thus representing a challenging benchmark for sequential multi-domain learning techniques. 
\noindent
\textbf{Accuracy Metrics.}
Both benchmarks are designed to address classification problems. Therefore, as common practice~\cite{mallya2018piggyback,mancini2018eccvw}, we report the accuracy for each domain and the average accuracy over the domains.
In addition, the score function $S$, as introduced in \cite{rebuffi2017learning}, is considered to jointly account for the $N$ domains. The test error $E_d$ of the model on the domain $d$ is compared to the test error of a baseline model $E_d^{\text{max}}$. The score is given by
%\begin{equation}\label{eq:S}
$S=\sum^{N}_{d=1}\alpha \max \{0,E_d^{\textrm{max}}-E_d\}^{2}$,
%\end{equation}
where $\alpha$ is a scaling parameter ensuring that the perfect score for each domain is 1000. The baseline error is given by doubling the error of 10 independent models fine-tuned for each domain. Importantly, this metric favors models with good performances over all domains, while penalizing those that are accurate only on few domains. %  peaked performances in few of them. % The S-score are reported for both benchmarks.

\noindent
\textbf{Complexity Metrics.}
Furthermore, since in this paper we argue that MDL methods should also be evaluated in terms of model complexity, we consider two other metrics which account for the number of network parameters and operations. First, following \cite{mallya2018piggyback,mancini2018eccvw}, we report the total number of parameters relative to the ones of the initial pre-trained model (counting all domains and excluding the classifiers). Note that, when computing the model size, we consider that all float numbers are encoded in 32 bits and switches in 1 bit only. Second, we propose to report the average number of floating-point operations (FLOP) over all the domains (including the pre-training domain $d=0$, i.e., ImageNet) relative to the number of operations of the initial pre-trained network. Interestingly, for the Budget-Aware Adapters, this ratio is also equal to the average number of parameters used at inference time for each individual domain, relative to the number of parameters of $\Psi_0$.

These complexity measures lead us to two variants of the score $S$: the score per parameter $S_P$ and the score per operation $S_O$. These two metrics are able to assess the trade-off between performance and model complexity.

\noindent
\textbf{Networks and training protocols.} Concerning the Visual Decathlon, we consider the Wide ResNet-28 \cite{zagoruyko2016wide} adopted in previous works \cite{mallya2018piggyback,mancini2018eccvw,rebuffi2017learning,rosenfeld2017incremental} and employ the same data pre-processing protocol. In term of hyper-parameters, we follow \cite{rebuffi2017learning} when pre-training on ImageNet. For other domains, we employ the hyper-parameters used in \cite{mallya2018piggyback}.

For the second benchmark, we use a ResNet-50 \cite{he2016deep}. %% , comparing our model with Piggyback \cite{mallya2018piggyback}, PackNet \cite{mallya2017packnet} and two baselines considering the network only as feature extractor (training only the domain-specific classifier) and individual networks separately fine-tuned on each domain.
Note that, since the performance of the method in \cite{mallya2017packnet} relies on the order of the domains, we report the performances for two orderings as in \cite{mallya2018piggyback}: starting from the model pre-trained on ImageNet, the first ($\rightarrow$) corresponds to CUBS-Cars-Flowers-WikiArt-Sketch, while the second ($\leftarrow$) corresponds to reversed order. We also followed the pre-processing, hyper-parameters and training schedule of \cite{mallya2018piggyback}, as we did in the Visual Decathlon Challenge.

We chose to use the same setting (network, pre-processing, and training schedules) employed by previous works seeking fairer analyses regarding the impact of the proposed approach.

\noindent
\textbf{Budget Constraints}
Even if our training procedure is formulated as a constrained optimization problem (see Equation~\ref{eq:opc}), the stochastic gradient descent algorithm we employ does not guarantee that all the constraints will be satisfied at the end of training. Therefore, in all our experiments, we check whether the final models respect the specified budget constraints. All the scores reported in this paper were obtained with models that respect the specified budget constraints, unless explicitly specified otherwise.

\input{tables/decathlon_flops.tex}
\begin{figure*}[t]
	\begin{center}
		%\fbox{\rule{0pt}{1in} \rule{0.9\linewidth}{0pt}}
	  \subfloat[][Computational complexity (FLOPs)]{\includegraphics[height=0.22\linewidth]{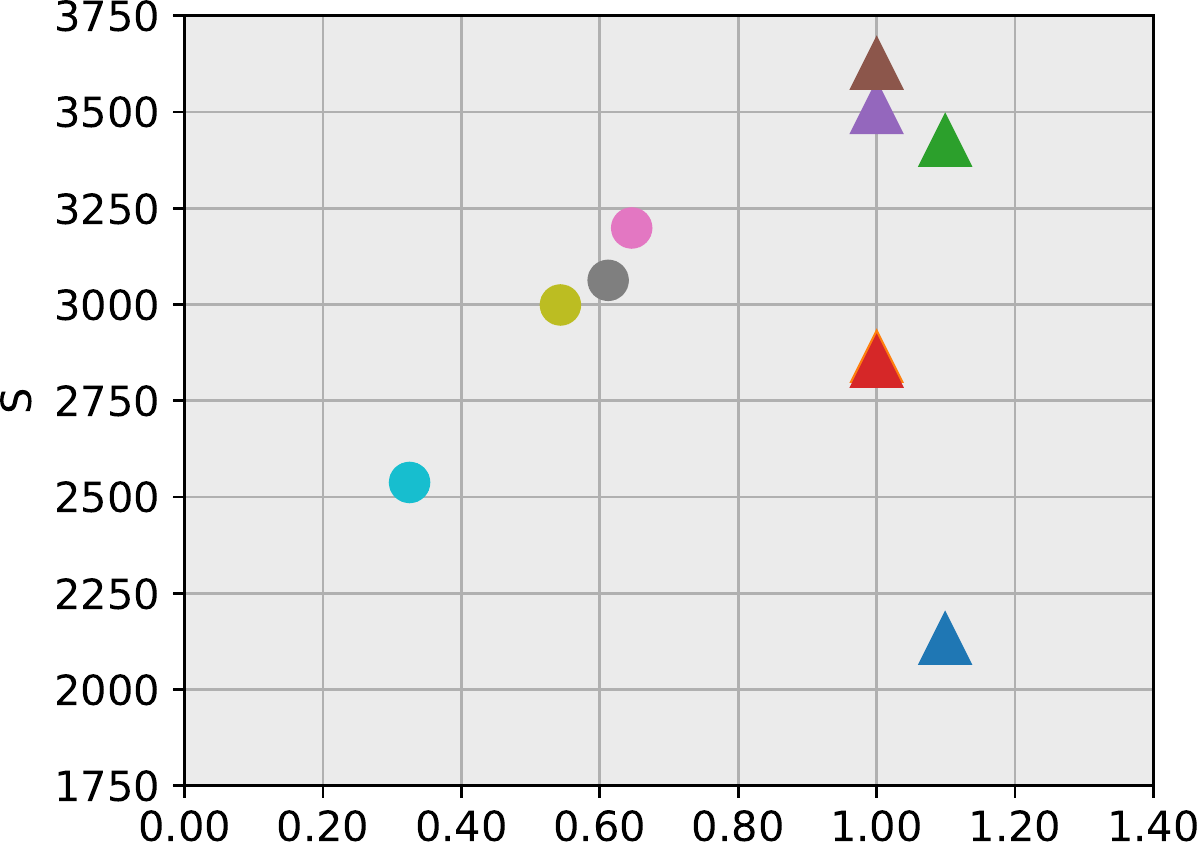}\label{fig:plotFlop}}
          %% \hspace{0.5cm}
	  \subfloat[][Network weight (Params)]{\includegraphics[height=0.22\linewidth]{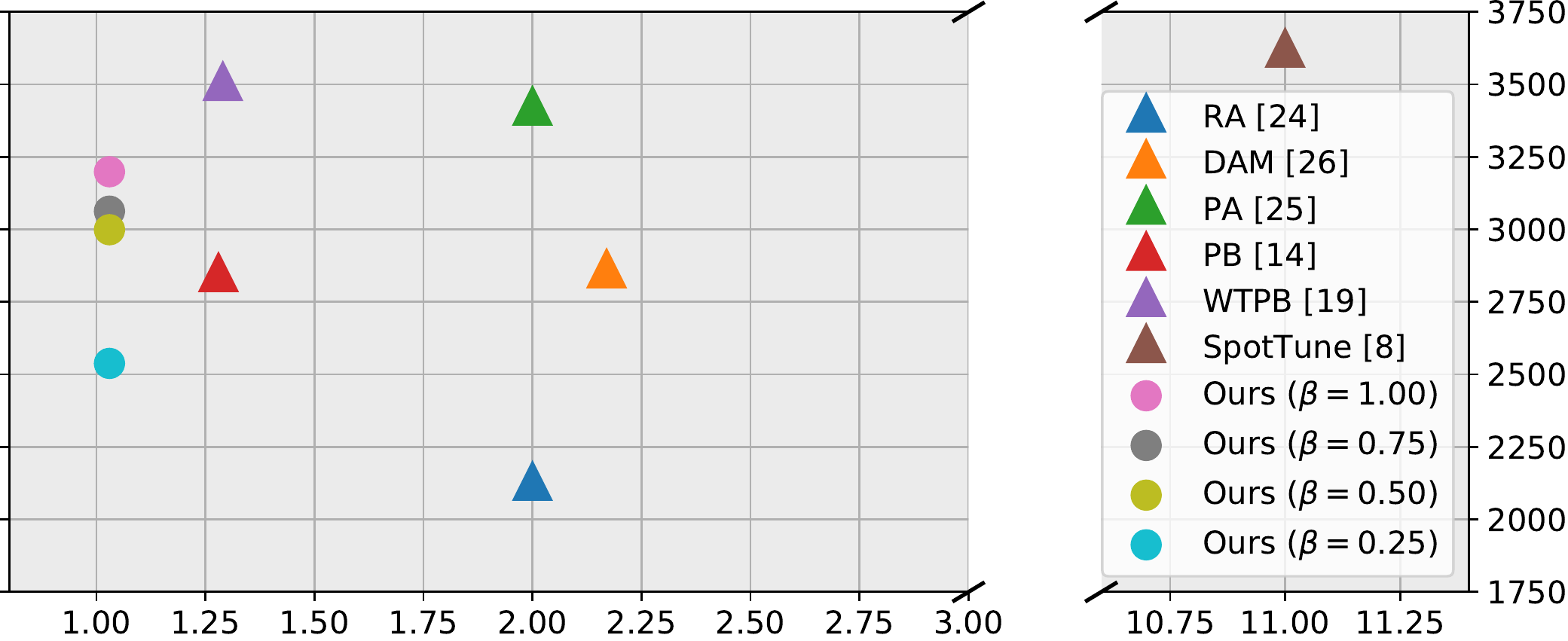}\label{fig:plotparamFig}}
	\end{center}
	\vspace{-0.5cm}
	\caption{Performance/Complexity Trade-Off on the visual decathlon challenge: the total score is displayed as a function of the two considered complexity metrics \emph{FLOPs} and \emph{Params}.}
	\label{fig:tradeoff}
\end{figure*}
\input{tables/image2sketch.tex}

\noindent
\textbf{Results on the Visual Decathlon Challenge.}
We first evaluate our methods on the Visual Decathlon Challenge. Results are reported in Table~\ref{tab:vdc-accuracy}. We report the \BA2 scores with respect to four different budgets $\beta \in \{0.25, 0.50, 0.75, 1.00\}$. We first observe that for most of the domains, \BA2 without budget constrains is competitive with state-of-the-art methods in terms of accuracy. Our method is the second best performing for three domains (GTSR, VGG-Flowers, and SVHN). In terms of score, among lightweight methods, only Parallel Adapters (PA) and Weight Transformations using Binary Masks (WTPB) perform better than ours. However, both methods require significantly more additional parameters to achieve these performances. Concerning the models where a budget constraint is considered, we observe that the scores still outperform RA~\cite{rebuffi2017learning}, DAM~\cite{rosenfeld2017incremental} and PB~\cite{mallya2018piggyback} in terms of score when targeting a budget of 50\% or 75\% of the initial network parameters, i.e., $\beta=0.50\textrm{ or }0.75$.

Interestingly, it can be seen in Table~\ref{tab:vdc-accuracy} that, when we impose a tighter budget to \BA2, the total number of parameters do not decrease. Indeed, all the parameters of the pre-trained network $\Psi_0$ are still required at testing time to handle the 10 domains. Only the number of parameters used for each domain and the number of floating-point operations are reduced. Therefore, we propose to complete this evaluation in order to further understand the performance/complexity trade-off achieved by each method. More precisely, we report the number of parameters and FLOPs in Table~\ref{tab:tradeoff}, and their corresponding scores. First, we observe that only \BA2 models report FLOPs lower than 1. In other words, only \BA2 provide models with fewer operations than the initial network $\Psi_0$. We see that \BA2 achieve the best performance in terms of $S_P$ when using 100\% budget. As mentioned above, the total number of parameters for the 10 domains do not decrease with a smaller budget. Consequently, smaller budgets obtain lower $S_P$ values. Nevertheless, the 75\% and 50\% models rank second and third, respectively.

Concerning the FLOP, only \BA2 return models with fewer floating-point operations than the initial network. As a consequence, \BA2 clearly outperforms other approaches in terms of $S_O$. 
In addition, we note that $S_O$ increases when using tighter budgets. It illustrates the potential of our approach in order to obtain a good performance/complexity trade-off.
Interestingly, even the models with $\beta=100\%$ report a FLOP value lower than 1 since convolutional channels can be dropped to adapt to each domain. Note that for all our models, the reported FLOP numbers are smaller than the specified budget. The reason for this is that we impose budget constraints independently to each convolutional layer in order to obtain a low memory footprint (see Subsection~\ref{sec:budget-conv-training}). Therefore, the average percentage of channels that are dropped can be smaller than the specified budget and, in fact, this is what we observed in ours models.% leading to models with lower FLOPs.

For better visualization, we illustrate in Figure~\ref{fig:tradeoff} the performance/complexity trade-off for each method on the Visual Decathlon Challenge.
More precisely, in Figure~\ref{fig:plotFlop}, we plot the score obtained as a function of the computation complexity in FLOPs. When comparing with other methods, we see that \BA2 lead to much lighter models that have, as a consequence, better performance/computation trade-offs. 
In Figure~\ref{fig:plotparamFig}, we report the obtained score as a function of the total number of parameters. \BA2 is the method that requires the lowest number of additional parameters to adapt to the 10 domains. Furthermore, this plot clearly show that our 100\% model has an interesting trade-off between the performance and the number of additional parameters. Note that WTPB~\cite{mancini2018eccvw} also obtained a good trade-off but stores, in total, 29\% more parameters (approximately $\times9$ per new domain). Interestingly, the best performing approach in terms of score, i.e., \emph{SpotTune}~\cite{guo2019spottune}, requires a much larger number of parameters that would restrict the use of this method when increasing the number of domains.

%% We first investigate the problem of learning multiple visually-diverse domains using our proposed \emph{budget-ConNets}. To this end, we employ the recent Visual Decathlon benchmark \cite{}. This benchmark is composed of 10 different well known datasets, from ImageNet\cite{}, to Aircraft \cite{} and German Traffic Signs \cite{}.

%% \begin{itemize}
%% 	\item Baseline: Piggyback;
%% 	\item Baseline: Quantized (Massi);
%% 	\item Piggyback + 1D masks;
%% 	\item Piggyback + 1D masks + 4x slim rates;
%% 	\item Piggyback + 1D masks + 4x slim rates + trainable kkt multiplier;
%% 	\item Piggyback + 1D masks + 10x slim rates + trainable kkt multiplier;
%% \end{itemize}
%% but we do not know what to do next nor what would be enough. The potential directions are:

%% \begin{itemize}
%% 	\item Add another set of domains;
%% 	\item Go for the "continuous" mask;
%% 	\item Try to apply the slimmable scheme to a problem with multi-modal input (possibly in the fusion);
%% \end{itemize}
%% but we did not decide what to do. The last option we still need to find a problem.

%% \begin{figure}[h]
%% 	\begin{center}
%% 		%\fbox{\rule{0pt}{1in} \rule{0.9\linewidth}{0pt}}
%% 		\includegraphics[width=0.99\linewidth]{figures/plotTO}
		
%% 	\end{center}
%%         \vspace{-0.7cm}
%% 	\caption{}
%% 	\label{fig:TO}
%% \end{figure}

\textbf{Results on the ImageNet-to-Sketch setting.}
We now compare our method with state-of-the-art approaches on the ImageNet-to-Sketch setting. Results are reported in Table~\ref{tabI2S}. First, we see that \BA2 achieve the second best score among methods that employ only a small number of additional parameters. Only WTPB~\cite{mancini2018eccvw} reports better scores at a higher cost in terms of additional parameters per domain. Interestingly, the 50\% and 75\% models report similar performances. Second, the $S_O$ and $S_P$ values clearly confirm the conclusions drawn on the first experiments on the Visual Decathlon Challenge. The four different \BA2 models outperform all the other methods in terms of $S_O$ and our model with 100\% budget slightly outperforms WTPB in terms of $S_P$.

\subsection{Ablation study of \textbf{BA}$^2$}

\begin{figure}[t]
    \centering
    \includegraphics[width=0.95\linewidth]{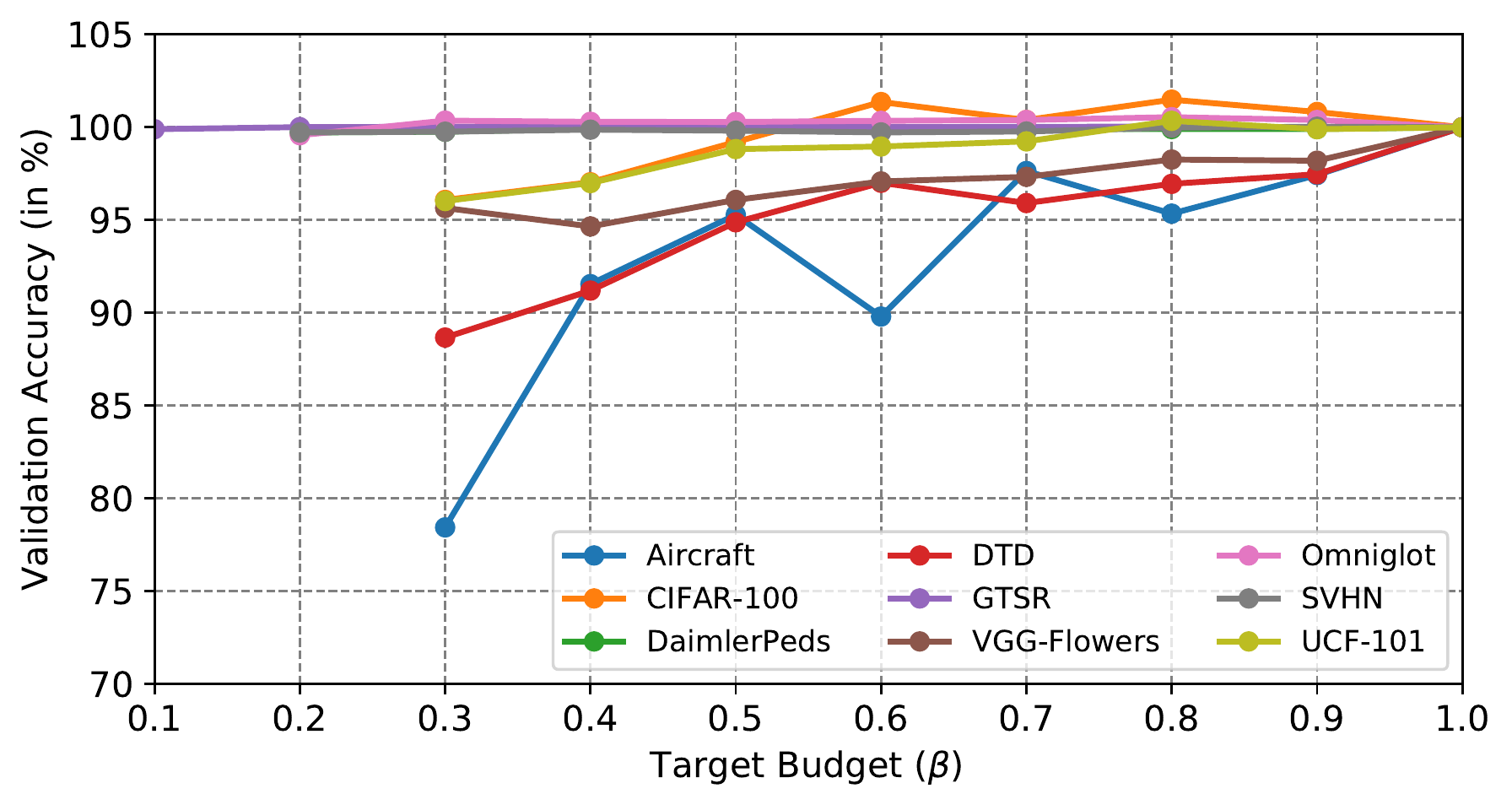}
    \vspace{-0.3cm}
    \caption{Relative accuracy drop compared to the 100\% model accuracy for the 10 domains of the Visual Decathlon challenge datset (validation set). Median score over 4 runs is reported. % and employ varying budget constraints from $\beta=10\%$ to $100\%$. 
    Missing points correspond to models where the budget constraint was not satisfied in the four runs.}
    \label{fig:decathlons-runs}
\end{figure}

In order to further understand the performance of \BA2, we propose to compare the drop in accuracy when imposing different budget constraints. In Figure~\ref{fig:decathlons-runs}, we perform an experiment on the Visual Decathlon Challenge with varying budgets $\beta=\{0.1, 0.2, \ldots, 1.0\}$. We display the accuracy drop relative to the performance of the model with 100\% budget. Because of the restrictions of the Challenge in terms of number of submissions (per day and in total), we report results on the validation set. To decrease the impact of the training stochasticity, we report the median performance over 4 runs.
As mentioned previously, the stochastic gradient descent algorithm we employ for training our model does not guarantee to provide a solution that respect the budget constraints. Therefore, in Figure~\ref{fig:decathlons-runs}, we display only points corresponding to models that satisfy the specified budget.
We first observe that for all the domains, our method returns models that respect the specified budget when the budget is greater than $30\%$. For some domains, we obtain models that respect even tighter budgets, such as the \emph{GTSR} dataset where we obtain a 10\%-budget model.

Interestingly, we notice that the domains where our models fail to respect the $20\%$ budget are the same in which the drop in performance is more clearly visible from the $100\%$ to $30\%$ budgets. Furthermore, we observe that the domains where the performance drop is small correspond to those where the $100\%$ model reaches excellent performance. For instance, in Table~\ref{tab:vdc-accuracy} the models for the \emph{GTSR} (traffic signs) and \emph{DPed} (pedestrians) datasets reach accuracy over $95\%$ with our $100\%$ model and do not significantly lose performance when imposing a tighter budget. Conversely, we observe that the \emph{DTD} and the \emph{aircraft} datasets, that show the largest performance drop, correspond also to the most challenging datasets according the accuracies of all methods reported in Table~\ref{tab:vdc-accuracy}.

\begin{figure}[ht]
	\begin{center}
		%\fbox{\rule{0pt}{1in} \rule{0.9\linewidth}{0pt}}
		\subfloat[][Cifar-10]{\includegraphics[height=0.36\linewidth]{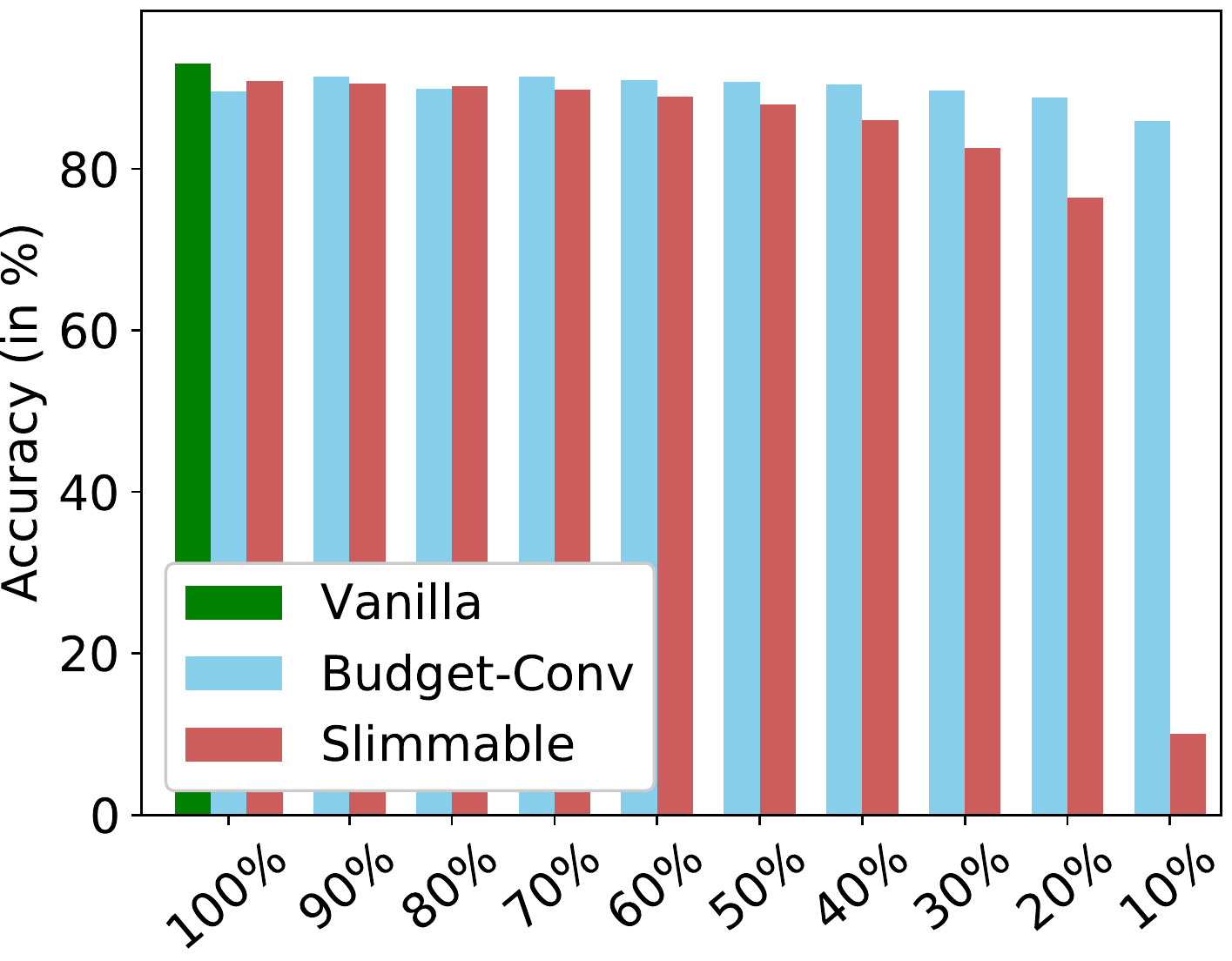}}
		\subfloat[][Cifar-100]{\includegraphics[height=0.36\linewidth]{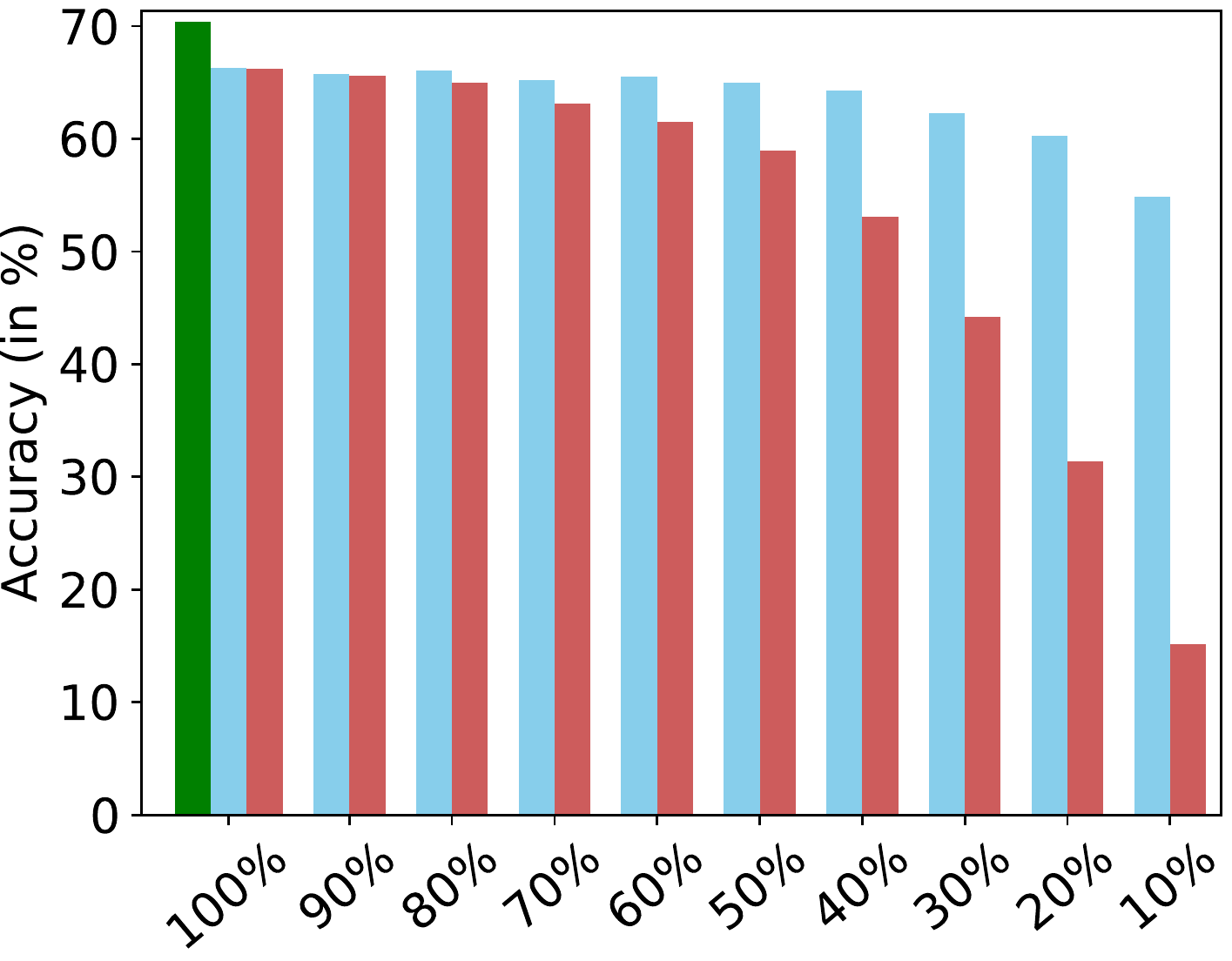}}
	\end{center}
        \vspace{-0.5cm}
	\caption{Single-domain classification with adaptive budget. \BA2 is compared with Slimmable Networks~\cite{yu2018slimmable}. }%\berriel{Slimmable=0.10 kind of collapsed; I have to double-check, but I think our budget=0.1 on CIFAR-100 is not satisfying the budget.}}
	\label{fig:cifar}
	
\end{figure}

\subsection{Evaluating \textbf{BA}$^2$ for single-domain problems}
\vspace{-0.2cm}
\label{exp:baa}
\noindent
In order to further demonstrate the effectiveness of our proposed \BA2, we perform experiments on a standard single-domain classification problem training a single ConvNet with different budget constraints.

\noindent
\textbf{Datasets and Experimental Protocol.}
We perform experiments on two well-known classification datasets: CIFAR-10 and CIFAR-100~\cite{krizhevsky2009learning}. In these experiments, we employ the following procedure. We first train a ResNet-50 architecture setting all the switch vectors to 1, leading to the initial model. We then employ different switches $\theta_s$ for each budget, but we share the convolution parameters $\theta_k$ among budgets. We fine-tune all the parameters optimizing Equation~\eqref{KKT} w.r.t. $\theta_k$ and all the different $\theta_s$ parameters jointly.
%% \end{itemize}
%%  Importantly, when using several budget containts, the second step is performed jointly for all the budgets in order to obtain shared $\theta_k$ parameters and, in this way limit the number of additional parameters.

\noindent
\textbf{Results.}
We compare our approach with the recently proposed Slimmable Networks~\cite{yu2018slimmable}. We consider this approach as this is the most closely related to our method in literature. In~\cite{yu2018slimmable}, different budgets are obtained by gradually dropping filters, imposing that filters dropped for a given budget are also dropped for lower budgets. Conversely, in \BA2 we do not impose any constraints between the switches at different budgets.
%For fair comparison, in this experiments, when training our budget-aware networks, we learn independent switches $\theta_s$ for each budget but share the convolution parameters $\theta_k$ over the different budgets. Results are reported in Fig.\ref{fig:cifar}.
It can be observed in Figure~\ref{fig:cifar} that, in the case of CIFAR-10, both models achieve an accuracy similar to a vanilla network trained without any budget constraint. On the more challenging CIFAR-100, both \BA2 and Slimmable Networks perform slightly worse than the vanilla network. Interestingly, \BA2 is able to maintain a constant accuracy on both datasets when decreasing the budget constraint up to 50\% whereas Slimmable Networks begins to perform poorly. The difference between the two methods becomes larger with tighter budgets, until 10\% where Slimmable Networks accuracy collapses. These experiments show that imposing constraints between budgets as in~\cite{yu2018slimmable} harms the performance and illustrate the potential of our approach even for single-domain problems.

%% file: tables/decathlon_flops.tex
\begin{table}[ht]

\centering
\resizebox{\columnwidth}{!}{%
\begin{tabular}{@{}l|ccc|cc@{}}
Method           & FLOP & Params  & Score & $S_O$ & $S_P$ \\ \toprule %\frac{\textrm{Score}}{\textrm{Params}}$ & $\frac{\textrm{Score}}{\textrm{FLOP}}$
Feature      & 1     & 1      & 544   & 544   & 544   \\
Finetune     & 1     & 10     & 2500  & 2500  & 250   \\
SpotTune     & 1     & 11     & 3612  & 3612  & 328   \\ \midrule
RA           & 1.099 & 2      & 2118  & 1926  & 1059  \\
DAM          & 1     & 2.17   & 2851  & 2851  & 1314  \\
PA           & 1.099 & 2      & 3412  & 3102  & 1706  \\
PB           & 1     & 1.28   & 2838  & 2838  & 2217  \\
WTPB         & 1     & 1.29   & \bf 3497  & 3497  & 2710  \\ \midrule
\BA2 (Ours) ($\beta=1.00$)    & 0.646 & \bf 1.03  & \underline{3199}  & 4952  & \bf 3106  \\
%\BA2 (Ours)                  & ?     & 1.03  & 2876  & ?     & 2792  \\
%\BA2 (Ours)                  & ?     & 1.03  & 2856  & ?     & 2773  \\
%\BA2 (Ours)                  & ?     & 1.03  & 2589  & ?     & 2514  \\
%\BA2 (Ours) (KKT)            & ?     & 1.03  & 3148  & ?     & 3056  \\
\BA2 (Ours) ($\beta=0.75$)    & 0.612 & \bf 1.03  & 3063  & 5005  & \underline{2974}  \\
\BA2 (Ours) ($\beta=0.50$)    & \underline{0.543} & \bf 1.03  & 2999  & \underline{5523}  & 2912  \\
\BA2 (Ours) ($\beta=0.25$)    & \bf 0.325 & \bf 1.03  & 2538  & \bf 7809  & 2464 
\end{tabular}%
}
\caption{Performance/Complexity trade-off comparison on the visual decathlon challenge.% \berriel{(*) These are computed based on the layer with minimum sparsity. The average is on Dropbox.}
\label{tab:tradeoff}}
\end{table}

%% file: tables/image2sketch.tex
\begin{table*}[ht]
\centering
\resizebox{0.92\textwidth}{!}{%
\begin{tabular}{@{}l|cc|cccccc|ccc@{}}
                                               & FLOP    & Params     & ImageNet  & CUBS    & Cars    & Flowers  & WikiArt  & Sketch & Score & $S_O$ & $S_P$  \\ \toprule
Classifier Only \cite{mallya2018piggyback}     & 1         & \bf 1         & 76.2      & 70.7    & 52.8    & 86.0     & 55.6     & 50.9   & 533  & 533  & 533  \\
Individual Networks \cite{mallya2018piggyback} & 1         & 6         & 76.2      & \underline{82.8}    & 91.8    & \bf 96.6     & \underline{75.6}     & \bf 80.8   & \bf 1500 & 1500 & 250  \\
SpotTune \cite{guo2019spottune}                & 1         & 7         & 76.2     & \bf 84.03   & \bf 92.40   & 96.34    & \bf 75.77    & \underline{80.2}  & \bf 1526 & 1526 & 218  \\ \midrule
PackNet $\rightarrow$ \cite{mallya2017packnet} & 1         & 1.10      & 75.7      & 80.4    & 86.1    & 93.0     & 69.4     & 76.2   & 732  & 732  & 665  \\
PackNet $\leftarrow$ \cite{mallya2017packnet}  & 1         & 1.10      & 75.7      & 71.4    & 80.0    & 90.6     & 70.3     & 78.7   & 620  & 620  & 534  \\
Piggyback \cite{mallya2018piggyback}           & 1         & 1.16      & 76.2      & 80.4    & 88.1    & 93.5     & 73.4     & 79.4   & 934  & 934  & 805  \\
Piggyback+BN \cite{mallya2018piggyback}        & 1         & 1.17      & 76.2      & 82.1    & 90.6    & 95.2     & 74.1     & 79.4   & 1184 & 1184 & 1012 \\
WTPB \cite{mancini2018eccvw}                   & 1         & 1.17      & 76.2      & 82.6    & 91.5    & \underline{96.5}     & 74.8     & \underline{80.2}   & 1430 & 1430 & \underline{1222} \\ \midrule
\BA2 (Ours) ($\beta=1.00$)                     & 0.700     & \underline{1.03}      & 76.2      & 81.19   & \underline{92.14}   & 95.74    & 72.32    & 79.28  & 1265 & 1807 & \bf 1228 \\
\BA2 (Ours) ($\beta=0.75$)                     & 0.600     & \underline{1.03}      & 76.2     & 79.44   & 90.62   & 94.44    & 70.92    & 79.38  & 1006 & 1677 & 977  \\
\BA2 (Ours) ($\beta=0.50$)                     & \underline{0.559}     & \underline{1.03}      & 76.2     & 79.34   & 90.80   & 94.91    & 70.61    & 78.28  & 1012 & \underline{1810} & 983  \\
\BA2 (Ours) ($\beta=0.25$)                     & \bf 0.375     & \underline{1.03}      & 76.2     & 78.01   & 88.15   & 93.19    & 67.99    & 77.85  & 755  & \bf 2013 & 733  
\end{tabular}%
}
\caption{State of the art comparison on the ImageNet-to-Sketch benchmark. Best model in bold, second best underlined.% \berriel{Maybe I should change everything to \{:.1f\}. *SpotTune's ImageNet numbers were not in their paper.}
\label{tabI2S}} 
\label{tab:resnet-ImageNet}
\end{table*}

%% file: sections/6_conclusions.tex
\vspace{-0.10cm}
\section{Conclusions}
\vspace{-0.2cm}
In this paper, we proposed to investigate the multi-domain learning problem with budget constraints. We propose to adapt a pre-trained network to each new domain with constraints on the network complexity. Our \emph{Budget-Aware Adapters} (\BA2) select the most relevant feature channels using trainable switch vectors. We impose constraints on the switches to obtain networks that respect to user-specified budget constraints. From an experimental point of view, \BA2 show performances competitive with state-of-the-art methods even with small budget values. As future works, we plan to extend our \emph{Budget-Aware Adapters} in order to be able do control the budget in a continuous fashion.

\noindent
\footnotesize{%
\vspace{-0.5cm}
\paragraph{\footnotesize{Acknowledgements}} This work was carried out under the ``Vision and Learning joint Laboratory'' between FBK and UNITN, and financed in part by the Coordena\c{c}\~ao de Aperfei\c{c}oamento de Pessoal de N\'ivel Superior -- Brasil (CAPES) -- Finance Code 001; CNPq, Brasil (311504/2017-5); and FAPES, Brasil (455/2019).
}

%% file: sections/7_supplementary.tex
\section{Additional training and testing details}
\label{sec:details}
In order to draw a fair comparison with other methods, we employed the same training schedule and hyperparameters of previous works. Here, we provide more details about the training and test procedures used in the three experiments: (i) Visual Decathlon Challenge, (ii) ImageNet-to-Sketch benchmark, and (iii) Single-Domain Classification.

\paragraph{Visual Decathlon challenge}
Following previous works~\cite{mallya2018piggyback, mancini2018eccvw, rebuffi2017learning, rosenfeld2017incremental}, we employ the Wide ResNet WRN-28-4-\textit{B}(3,3), i.e., 28 convolutional layers with widening factor of 4 and the original ``basic'' block (2 convolutions using $3\times3$ kernel size). As in~\cite{rebuffi2017learning, mancini2018eccvw}, random crops of $64 \times 64$ pixels are used to feed the network during training. The optimizer parameters are set following~\cite{mallya2018piggyback, mancini2018eccvw}, where SGD with momentum is employed for the classifier and Adam for the rest of the architecture with initial learning rates of 0.001 and 0.0001, respectively. The model is trained with batch size of 32 for 60 epochs; after 45 epochs, the learning rates are decayed by a factor of 10. The real-valued switches ($\tilde{s}_c$) are initialized with a value of 0.001. In addition to random cropping, in regards to training data augmentation, horizontal mirroring is also applied with a 50\% probability, except for four datasets (DTD, Omniglot, SVHN, and GTSR) on which it could be either harmful or useless. During training, the models for all the domains are initialized using the ImageNet pre-trained weights and these weights are kept fixed, i.e., only the domain-specific parameters ($\tilde{s}_c$, Batch Normalization and classifiers) are learned. At test time, we follow the procedure proposed in \cite{krizhevsky2012imagenet} and used in \cite{mancini2018eccvw}. We employ a ten-crop strategy for the datasets in which horizontal mirroring was applied and five-crop otherwise. In the five-crop strategy, a crop is performed on each corner of the image in addition to a central one; the ten-crop adds horizontally mirrored versions of each one of the five crops. The final prediction is based on the average of the predictions over all the crops.

\paragraph{ImageNet-to-Sketch benchmark}
We use the ResNet-50 as in~\cite{mallya2018piggyback, mancini2018eccvw} feeding a random crop of $224 \times 224$ pixels after resizing the images to $256 \times 256$ pixels. In addition to random cropping, horizontal mirroring is applied to all datasets during training. The networks are initially trained using the same schedule as in~\cite{mallya2018piggyback, mancini2018eccvw}, i.e., 30 epochs with a learning rate drop (with a factor of 10) after 15 epochs. For the lowest budget ($\beta=0.25$), we add another learning rate drop at 30 epochs and train for additional 15 epochs in order to fully satisfy the constraints.
%However, the constraints of the lowest budget ($\beta=0.25$) was not being satisfied (by a small margin) and, to address this, we added another learning rate drop at 30 epochs and trained for additional 15 epochs. Almost no improvement could be noticed on the other budgets, but with this change the lowest budget is almost always satisfied as well. \berriel{Should I keep this difference of schedule here?} 
All the other steps are performed as in the Visual Decathlon Challenge.

\paragraph{Single-domain classification}
In these experiments, we use the same variant of the ResNet proposed in the original Residual Networks~\cite{he2016deep} paper for the CIFAR-10 dataset with $n=9$, which results in a ResNet with 56 layers. First, we train the baseline model, which is the model without \textit{switches}. Since we are interested in single-domain learning for this experiment, we do not need to freeze the weights as in the multi-domain experiments. Therefore, we jointly train the switches and the weights using the baseline pre-trained weights as initialization. In these cases, we use SGD with momentum (initial learning rate of $0.1$) for both the baseline model and the joint training. Concerning data augmentation, we use random crops of $32 \times 32$ pixels with 4 pixel padding and horizontal mirroring with 50\% probability. The same setting is used for the CIFAR-100.

\section{Results}
\label{sec:denseNet}
\input{tables/densenet.tex}
\paragraph{ImageNet-to-Sketch}
In the paper, we report the results on the ImageNet-to-Sketch benchmark using the ResNet-50 architecture. In addition to this model, results using the DenseNet-121 are also reported in several works~\cite{mallya2018piggyback,mallya2017packnet,mancini2018eccvw}. For that reason, we also provide these results in Table~\ref{tab:densenet-i2s}. First, we see that our method achieves the best scores in two domains and the second best in three other domains. Interestingly, in the \emph{Cars} domain, our method with $\beta = 0.75$ is the second best model (only after our method with $\beta = 1.00$), achieving results better than all the other methods using only 57.8\% of the FLOP, on average. Concerning the number parameters, our approach is the second best in terms of Params. Indeed the number of batch normalization and $1 \times 1$ convolutions parameters in the DenseNet-121 model with respect to the total number of parameters is higher than in the ResNet-50 model. Nevertheless, \BA2 is still the only one with FLOP less than 1.
Finally, it can be noted that our method with $\beta = 1.00$ still achieves a good trade-off between performance and complexity.
%As can be seen \berriel{I have to write something here. WikiArt numbers are not that good, again. If I choose another epoch instead of the last one, than we can be better, ofc.}

%% file: tables/densenet.tex
\begin{table*}[ht]
\centering
\resizebox{\textwidth}{!}{%
\begin{tabular}{@{}l|cc|cccccc|ccc@{}}
                                               & FLOP    & Params     & ImageNet  & CUBS    & Cars    & Flowers  & WikiArt  & Sketch & Score & $S_O$ & $S_P$  \\ \toprule
Classifier Only \cite{mallya2018piggyback}     & 1      & 1  & 74.4  & 73.5   & 56.8   & 83.4  & 54.9  & 53.1   & 328   & 328   & 328 \\
Individual Networks \cite{mallya2018piggyback} & 1      & 6  & 74.4  & 81.7   & 91.4   & 96.5  & 76.4  & 80.5   & 1500  & 1500  & 250 \\
\midrule
PackNet $\rightarrow$ \cite{mallya2017packnet} & 1      & \bf 1.11  & 74.4  & 80.7   & 84.7   & 91.1  & 66.3  & 74.7   & 691   & 691   & 623 \\
PackNet $\leftarrow$ \cite{mallya2017packnet}  & 1      & \bf 1.11  & 74.4  & 69.6   & 77.9   & 91.5  & 69.2  & 78.9   & 610   & 610   & 550 \\
Piggyback \cite{mallya2018piggyback}           & 1      & 1.15  & 74.4  & 79.7   & 87.2   & 94.3  & 72.0  & \bf 80.0   & 951   & 951   & 827 \\
Piggyback+BN \cite{mallya2018piggyback}        & 1      & 1.21  & 74.4  & 81.4   & 90.1   & 95.5  & \underline{73.9}  & 79.1   & 1215  & 1215  & 1004 \\
WTPB \cite{mancini2018eccvw}                   & 1      & 1.21  & 74.4  & \underline{81.7}   & 91.6   & \bf 96.9  & \bf 75.7  & 79.8   & \bf 1540  & 1540  & \bf 1268 \\
\midrule
\BA2 (Ours) ($\beta=1.00$)                     & 0.687  & \underline{1.17}  & 74.4 & \bf 82.4  & \bf 92.9  & \underline{96.0}  & 71.5  & \underline{79.9}  & \underline{1440}  & \underline{2096}  & \underline{1230} \\
\BA2 (Ours) ($\beta=0.75$)                     & 0.578  & \underline{1.17}  & 74.4 & 81.2  & \underline{91.9}  & 94.9  & 68.9  & \underline{79.9}  & 1193  & 2064  & 1019 \\
\BA2 (Ours) ($\beta=0.50$)                     & 0.543  & \underline{1.17}  & 74.4 & 78.2  & 89.2  & 95.0  & 66.2  & 78.8  & 925   & 1703  & 790 \\
\BA2 (Ours) ($\beta=0.25$)                     & \bf 0.375  & \underline{1.17}  & 74.4 & 76.2* & 88.4* & 94.7* & 67.9* & 78.4  & 840   & \bf 2240  & 717  
\end{tabular}%
}
\caption{State-of-the-art comparison on the ImageNet-to-Sketch benchmark using DenseNet-121 architecture. (*) Even though the average sparsities are greater than 75\%, these models did not satisfy the constraint for every single layer.}
\label{tab:densenet-i2s}
\end{table*}

%% file: main_arxiv.bbl
\begin{thebibliography}{10}\itemsep=-1pt

\bibitem{aljundi2018memory}
Rahaf Aljundi, Francesca Babiloni, Mohamed Elhoseiny, Marcus Rohrbach, and
  Tinne Tuytelaars.
\newblock {Memory Aware Synapses: Learning what (not) to forget}.
\newblock In {\em European Conference on Computer Vision (ECCV)}, 2018.

\bibitem{BendaleB16}
Abhijit Bendale and Terrance~E. Boult.
\newblock {Towards Open Set Deep Networks}.
\newblock In {\em Conference on Computer Vision and Pattern Recognition
  (CVPR)}, 2016.

\bibitem{bilen2016dynamic}
Hakan Bilen, Basura Fernando, Efstratios Gavves, Andrea Vedaldi, and Stephen
  Gould.
\newblock {Dynamic Image Networks for Action Recognition}.
\newblock In {\em Conference on Computer Vision and Pattern Recognition
  (CVPR)}, 2016.

\bibitem{chen2018deeplab}
{Liang-Chieh} Chen, George Papandreou, Iasonas Kokkinos, Kevin Murphy, and
  Alan~L Yuille.
\newblock {DeepLab: Semantic Image Segmentation with Deep Convolutional Nets,
  Atrous Convolution, and Fully Connected CRFs}.
\newblock {\em IEEE Transactions on Pattern Analysis and Machine Intelligence},
  40(4):834--848, 2018.

\bibitem{cimpoi2014describing}
Mircea Cimpoi, Subhransu Maji, Iasonas Kokkinos, Sammy Mohamed, and Andrea
  Vedaldi.
\newblock {Describing Textures in the Wild}.
\newblock In {\em Conference on Computer Vision and Pattern Recognition
  (CVPR)}, 2014.

\bibitem{eitz2012humans}
Mathias Eitz, James Hays, and Marc Alexa.
\newblock {How Do Humans Sketch Objects?}
\newblock {\em ACM Transactions on Graphics}, 31(4):44:1--44:10, 2012.

\bibitem{girshick2014rich}
Ross Girshick, Jeff Donahue, Trevor Darrell, and Jitendra Malik.
\newblock {Rich feature hierarchies for accurate object detection and semantic
  segmentation}.
\newblock In {\em Conference on Computer Vision and Pattern Recognition
  (CVPR)}, 2014.

\bibitem{guo2019spottune}
Yunhui Guo, Honghui Shi, Abhishek Kumar, Kristen Grauman, Tajana Rosing, and
  Rogerio Feris.
\newblock {SpotTune: Transfer Learning through Adaptive Fine-tuning}.
\newblock In {\em Conference on Computer Vision and Pattern Recognition
  (CVPR)}, 2019.

\bibitem{he2016deep}
Kaiming He, Xiangyu Zhang, Shaoqing Ren, and Jian Sun.
\newblock {Deep Residual Learning for Image Recognition}.
\newblock In {\em Conference on Computer Vision and Pattern Recognition
  (CVPR)}, 2016.

\bibitem{kaiser2018fast}
{\L}ukasz Kaiser, Aurko Roy, Ashish Vaswani, Niki Parmar, Samy Bengio, Jakob
  Uszkoreit, and Noam Shazeer.
\newblock {Fast Decoding in Sequence Models Using Discrete Latent Variables}.
\newblock In {\em International Conference on Machine Learning (ICML)}, 2018.

\bibitem{kirkpatrick2017overcoming}
James Kirkpatrick, Razvan Pascanu, Neil Rabinowitz, Joel Veness, Guillaume
  Desjardins, Andrei~A Rusu, Kieran Milan, John Quan, Tiago Ramalho, Agnieszka
  Grabska-Barwinska, et~al.
\newblock {Overcoming catastrophic forgetting in neural networks}.
\newblock {\em Proceedings of the National Academy of Sciences},
  114(13):3521--3526, 2017.

\bibitem{krause20133d}
Jonathan Krause, Michael Stark, Jia Deng, and Li Fei-Fei.
\newblock {3D Object Representations for Fine-Grained Categorization}.
\newblock In {\em International Conference on Computer Vision Workshops
  (ICCVW)}, 2013.

\bibitem{krizhevsky2009learning}
Alex Krizhevsky.
\newblock {Learning multiple layers of features from tiny images}.
\newblock Master's thesis, University of Toronto, 2009.

\bibitem{krizhevsky2012imagenet}
Alex Krizhevsky, Ilya Sutskever, and Geoffrey~E. Hinton.
\newblock {ImageNet Classification with Deep Convolutional Neural Networks}.
\newblock In {\em Advances in Neural Information Processing Systems (NeurIPS)},
  2012.

\bibitem{lake2015human}
Brenden~M. Lake, Ruslan Salakhutdinov, and Joshua~B. Tenenbaum.
\newblock {Human-level concept learning through probabilistic program
  induction}.
\newblock {\em Science}, 350(6266):1332--1338, 2015.

\bibitem{li2017learning}
Zhizhong Li and Derek Hoiem.
\newblock {Learning without Forgetting}.
\newblock {\em IEEE Transactions on Pattern Analysis and Machine Intelligence},
  40(12):2935--2947, 2017.

\bibitem{maji2013fine}
Subhransu Maji, Esa Rahtu, Juho Kannala, Matthew Blaschko, and Andrea Vedaldi.
\newblock {Fine-Grained Visual Classification of Aircraft}.
\newblock {\em arXiv preprint arXiv:1306.5151}, 2013.

\bibitem{mallya2018piggyback}
Arun Mallya, Dillon Davis, and Svetlana Lazebnik.
\newblock {Piggyback: Adapting a Single Network to Multiple Tasks by Learning
  to Mask Weights}.
\newblock In {\em European Conference on Computer Vision (ECCV)}, 2018.

\bibitem{mallya2017packnet}
Arun Mallya and Svetlana Lazebnik.
\newblock {PackNet: Adding Multiple Tasks to a Single Network by Iterative
  Pruning}.
\newblock In {\em Conference on Computer Vision and Pattern Recognition
  (CVPR)}, 2018.

\bibitem{mancini2018eccvw}
Massimiliano Mancini, Elisa Ricci, Barbara Caputo, and Samuel~Rota Bul{\`o}.
\newblock {Adding New Tasks to a Single Network with Weight Transformations
  using Binary Masks}.
\newblock In {\em European Conference on Computer Vision Workshops (ECCVW)},
  2018.

\bibitem{munder2006experimental}
Stefan Munder and Dariu~M. Gavrila.
\newblock {An Experimental Study on Pedestrian Classification}.
\newblock {\em IEEE Transactions on Pattern Analysis and Machine Intelligence},
  28(11):1863--1868, 2006.

\bibitem{netzer2011reading}
Yuval Netzer, Tao Wang, Adam Coates, Alessandro Bissacco, Bo Wu, and Andrew~Y.
  Ng.
\newblock {Reading Digits in Natural Images with Unsupervised Feature
  Learning}.
\newblock In {\em NeurIPS Workshop on Deep Learning and Unsupervised Feature
  Learning}, 2011.

\bibitem{nilsback2008automated}
Maria-Elena Nilsback and Andrew Zisserman.
\newblock {Automated Flower Classification over a Large Number of Classes}.
\newblock In {\em Indian Conference on Computer Vision, Graphics \& Image
  Processing}, 2008.

\bibitem{RebuffiKSL17}
Sylvestre{-}Alvise Rebuffi, Alexander Kolesnikov, Georg Sperl, and Christoph~H.
  Lampert.
\newblock {iCaRL: Incremental Classifier and Representation Learning}.
\newblock In {\em Conference on Computer Vision and Pattern Recognition
  (CVPR)}, 2017.

\bibitem{rebuffi2017learning}
Sylvestre-Alvise Rebuffi, Hakan Bilen, and Andrea Vedaldi.
\newblock {Learning multiple visual domains with residual adapters}.
\newblock In {\em Advances in Neural Information Processing Systems (NeurIPS)},
  2017.

\bibitem{rebuffi2018efficient}
Sylvestre-Alvise Rebuffi, Hakan Bilen, and Andrea Vedaldi.
\newblock {Efficient parametrization of multi-domain deep neural networks}.
\newblock In {\em Conference on Computer Vision and Pattern Recognition
  (CVPR)}, 2018.

\bibitem{rosenfeld2017incremental}
Amir Rosenfeld and John~K. Tsotsos.
\newblock {Incremental Learning Through Deep Adaptation}.
\newblock {\em IEEE Transactions on Pattern Analysis and Machine Intelligence},
  2018.
\newblock Early Access.

\bibitem{russakovsky2015imagenet}
Olga Russakovsky, Jia Deng, Hao Su, Jonathan Krause, Sanjeev Satheesh, Sean Ma,
  Zhiheng Huang, Andrej Karpathy, Aditya Khosla, Michael Bernstein, et~al.
\newblock {ImageNet Large Scale Visual Recognition Challenge}.
\newblock {\em International Journal of Computer Vision}, 115(3):211--252,
  2015.

\bibitem{rusu2016progressive}
Andrei~A. Rusu, Neil~C. Rabinowitz, Guillaume Desjardins, Hubert Soyer, James
  Kirkpatrick, Koray Kavukcuoglu, Razvan Pascanu, and Raia Hadsell.
\newblock {Progressive Neural Networks}.
\newblock {\em arXiv preprint arXiv:1606.04671}, 2016.

\bibitem{saleh2015large}
Babak Saleh and Ahmed Elgammal.
\newblock {Large-scale Classification of Fine-Art Paintings: Learning The Right
  Metric on The Right Feature}.
\newblock {\em International Journal for Digital Art History}, 2016.

\bibitem{mobilenet2018cvpr}
Mark Sandler, Andrew Howard, Menglong Zhu, Andrey Zhmoginov, and Liang-Chieh
  Chen.
\newblock {MobileNetV2: Inverted Residuals and Linear Bottlenecks}.
\newblock In {\em Conference on Computer Vision and Pattern Recognition
  (CVPR)}, 2018.

\bibitem{soomro2012ucf101}
Khurram Soomro, Amir~Roshan Zamir, and Mubarak Shah.
\newblock {UCF101: A Dataset of 101 Human Actions Classes From Videos in The
  Wild}.
\newblock {\em arXiv preprint arXiv:1212.0402}, 2012.

\bibitem{stallkamp2012man}
Johannes Stallkamp, Marc Schlipsing, Jan Salmen, and Christian Igel.
\newblock {Man vs. computer: Benchmarking machine learning algorithms for
  traffic sign recognition}.
\newblock {\em Neural Networks}, 32:323--332, 2012.

\bibitem{wang2018skipnet}
Xin Wang, Fisher Yu, Zi-Yi Dou, Trevor Darrell, and Joseph~E. Gonzalez.
\newblock {SkipNet: Learning Dynamic Routing in Convolutional Networks}.
\newblock In {\em European Conference on Computer Vision (ECCV)}, 2018.

\bibitem{wah2011caltech}
Peter Welinder, Steve Branson, Takeshi Mita, Catherine Wah, Florian Schroff,
  Serge Belongie, and Pietro Perona.
\newblock {Caltech-UCSD Birds 200}.
\newblock Technical Report CNS-TR-2010-001, California Institute of Technology,
  2010.

\bibitem{wu2018blockdrop}
Zuxuan Wu, Tushar Nagarajan, Abhishek Kumar, Steven Rennie, Larry~S Davis,
  Kristen Grauman, and Rogerio Feris.
\newblock {BlockDrop: Dynamic Inference Paths in Residual Networks}.
\newblock In {\em Conference on Computer Vision and Pattern Recognition
  (CVPR)}, 2018.

\bibitem{xu2017multi}
Dan Xu, Elisa Ricci, Wanli Ouyang, Xiaogang Wang, and Nicu Sebe.
\newblock {Multi-Scale Continuous CRFs as Sequential Deep Networks for
  Monocular Depth Estimation}.
\newblock In {\em Conference on Computer Vision and Pattern Recognition
  (CVPR)}, 2017.

\bibitem{yu2018slimmable}
Jiahui Yu, Linjie Yang, Ning Xu, Jianchao Yang, and Thomas Huang.
\newblock {Slimmable Neural Networks}.
\newblock In {\em International Conference on Learning Representations (ICLR)},
  2019.

\bibitem{zagoruyko2016wide}
Sergey Zagoruyko and Nikos Komodakis.
\newblock {Wide Residual Networks}.
\newblock In {\em British Machine Vision Conference (BMVC)}, 2016.

\end{thebibliography}
